\documentclass[lettersize,journal]{IEEEtran}
\usepackage{amsmath,amsfonts}
\usepackage{algorithmic}
\usepackage{algorithm}
\usepackage{array}
\usepackage[caption=false,font=normalsize,labelfont=sf,textfont=sf]{subfig}
\usepackage{textcomp}
\usepackage{stfloats}
\usepackage{url}
\usepackage{verbatim}
\usepackage{graphicx}
\usepackage{cite}
\usepackage{xcolor}
\usepackage[normalem]{ulem}
\usepackage{array}
\usepackage{multirow}
\usepackage{booktabs}
\usepackage{hyperref}
\hypersetup{hypertex=true,
colorlinks=true,
linkcolor=blue,
anchorcolor=blue,
citecolor=blue}
\usepackage{makecell} 
\usepackage[nameinlink]{cleveref}
\begin{document}

\title{A Cosine Network for Image Super-Resolution}

\author{Chunwei Tian, \emph{Member}, \emph{IEEE},
        Chengyuan Zhang,
        Bob Zhang, \emph{Senior Member}, \emph{IEEE},
        Zhiwu Li, \emph{Fellow}, \emph{IEEE},
        C. L. Philip Chen, \emph{Life Fellow}, \emph{IEEE},
        David Zhang, \emph{Life Fellow}, \emph{IEEE}
        \\
\thanks{This work was supported in part by the National Natural Science Foundation of China under Grant 62576123 and 92267203, in part by the University of Macau (MYRG-GRG2024-00205-FST-UMDF) and the Science and Technology Development Fund(FDCT), Macao S.A.R under Grant 0028/2023/RIA1, and in part by the Program for Guangdong Introducing Innovative and Entrepreneurial Teams (2019ZT08X214). (\textit{Corresponding author: Chengyuan Zhang; Bob Zhang.})}
\thanks{Chunwei Tian is with the School of Computer Science and Technology, Harbin Institute of Technology, Harbin 150001, China. (Email:chunweitian@hit.edu.cn)}
\thanks{Chengyuan Zhang is with the College of Computer Science and Electronic Engineering, Hunan University, Changsha 410082, China. (Email: cyzhangcse@hun.edu.cn).}
\thanks{Bob Zhang is with the PAMI Research Group, Dept. of Computer and Information Science and the Centre for Artificial Intelligence and Robotics, Institute of Collaborative Innovation, University of Macau, Taipa 999078, Macau SAR, China. (Email:bobzhang@um.edu.mo).}
\thanks{Zhiwu Li is with the School of Elector-Mechanical Engineering, Xidian University, Xi’an 710071, China. (e-mail: zhwli@xidian.edu.cn).}
\thanks{C. L. Philip Chen is with the School of Computer Science and Engineering, South China University of Technology and Pazhou Lab, Guangzhou 510641, 510335, China. (e-mail: philip.chen@ieee.org).}
\thanks{David Zhang is with the School of Data Science, Chinese University of Hong Kong (Shenzhen), Shenzhen 518172, China, and also with Shenzhen Institute of Artificial Intelligence and Robotics for Society, Shenzhen 518172, China. (e-mail: davidzhang@cuhk.edu.cn).}
}

\markboth{
Journal of \LaTeX\ Class Files,~Vol.~14, No.~8, August~2021
}
{
Shell \MakeLowercase{\textit{et al.}}: A Sample Article Using IEEEtran.cls for IEEE Journals
}

\maketitle

\begin{abstract}
Deep convolutional neural networks can use hierarchical information to progressively extract structural information to recover high-quality images. However, preserving the effectiveness of the obtained structural information is important in image super-resolution. In this paper, we propose a cosine network for image super-resolution (CSRNet) by improving a network architecture and optimizing the training strategy. To extract complementary homologous structural information, odd and even heterogeneous blocks are designed to enlarge the architectural differences and improve the performance of image super-resolution. Combining linear and non-linear structural information can overcome the drawback of homologous information and enhance the robustness of the obtained structural information in image super-resolution. Taking into account the local minimum of gradient descent, a cosine annealing mechanism is used to optimize the training procedure by performing warm restarts and adjusting the learning rate. Experimental results illustrate that the proposed CSRNet is competitive with state-of-the-art methods in image super-resolution.
\end{abstract}


\section{Introduction}
\IEEEPARstart{S}{ingle} image super-resolution (SISR) can use machine learning methods to restore high-resolution (HR) images from low-resolution (LR) images, which is important for high-level vision, e.g., medical image classification \cite{ma2020pathsrgan}, object recognition \cite{sajjadi2017enhancenet}, remote sensing satellite image recognition \cite{zhang2020remote}, etc. Due to the ill-posed nature of the super-resolution task, detailed information is easily lost during image restoration. To overcome this drawback, traditional machine learning methods have three categories: super-resolution methods based on interpolation, reconstruction, and learning. The first method directly uses interpolation to enlarge the image \cite{zhou2012interpolation}, but it tends to lose details when the up-scaling factor is large.  {The second method analyzes relations among multiple low-resolution images of the same scene}, then it relies on prior information in the reconstruction process to obtain a high-resolution image \cite{park2003super}.  For instance, Tasi et al. \cite{tsai1984multiframe} exploited the Fourier transform to recover HR images from multiple LR images. Although it can effectively deal with image super-resolution, it suffers from the adverse effects of noise and blur in the real world in image super-resolution \cite{nguyen1999blind}. Rhee et al. \cite{rhee1999discrete} simultaneously used generalized regularization in the spatial domain and discrete cosine transform to improve reconstruction efficiency and enhance stability for different scenes. Demirel et al. \cite{demirel2010image} presented two mixed wavelet methods, i.e., discrete and stationary wavelets to enhance edge information to predict high-definition images. Alternatively, horizontal and vertical shifts in the gradient algorithm are used to improve super-resolution \cite{panagiotopoulou2007super}. To improve adaptability for different conditions, Wang et al. simultaneously used co-occurrence prior and reconstruction constraints to implement a probabilistic framework for image super-resolution \cite{wang2005patch}. Yang et al. \cite{yang2008image} proposed a sparse coding method to encode patches to implement a dictionary learning for restoring more detailed information of high-quality images.	Alternatively, Chang et al. \cite{chang2004super} exploited locally linear embedding to estimate an HR image patch from an LR image patch to implement image SR. Although the methods above can effectively improve the resolutions of given LR images, they still face the following challenges of manually setting parameters and complex optimal algorithms. Although learning-based methods achieve excellent performance, they often rely on complex algorithms and manual tuning parameters, which not only reduces flexibility but also makes it difficult to improve SISR efficiency and optimize model performance. 

Due to flexible architectures, convolutional neural networks (CNNs) have been widely applied in low-level vision tasks, i.e., SISR. A network consisting of three stacked convolutional layers can be used to obtain a high-definition image from a low-resolution image in a pixel mapping way for image super-resolution \cite{dong2015image}. Although the proposed methods mentioned were very effective, they may suffer from a challenge between SR performance and network depth. To overcome this issue, deeper networks with residual learning operations are developed \cite{kim2016accurate,kim2016deeply}. For instance, Kim et al. exploited deep networks with a residual learning operation to extract more accurate structural information for improving the quality of recovered images \cite{kim2016accurate}. Alternatively, recursive-supervision and skip-connection were embedded into a CNN to ease training difficulty in image super-resolution \cite{kim2016deeply}. To reduce parameters, residual learning operations were used in a global and local network to reduce the difficulty of training in image super-resolution \cite{tai2017image}. To reduce complexity, low-resolution images were directly used as inputs of a deep network to reduce computational costs \cite{dong2016accelerating}. Inspired by that, residual learning operations are repeatedly used in a CNN to strengthen structural information to acquire more high-quality images \cite{lim2017enhanced}. To enhance internal relations of different layers, Zhang et al. \cite{zhang2018image} used a channel mechanism to enhance channel relations to extract salient structural information in image super-resolution. Although these SR methods are effective in restoring high-quality images, most of them rely on deep architectures to achieve a better effect in image super-resolution, which cannot preserve the effectiveness of each layer in SISR. 

We design a cosine network for SISR, termed CSRNet, in this paper, which has improvements in enhancing network architecture and optimizing training strategy. In terms of designing new network architecture, odd and even heterogeneous blocks are designed to enlarge the architectural differences and obtain more homologous structural information for image super-resolution. Combining heterogeneous structural information is complementary to homologous structural information to pursue better SR performance. Taking into account the local minimum of gradient descent, a cosine annealing mechanism is used to optimize training parameters for image SR. Experimental results are competitive in contrast to popular SR methods. 

The contributions of the proposed CSRNet are summarized as follows.  

(1) A heterogeneous network can extract complementary homologous structural information by designing odd and even heterogeneous blocks for image super-resolution

(2) Extracting heterogeneous structural information can cooperate with homologous structural information to improve the performance of image super-resolution.

(3) A cosine annealing mechanism can be used to optimize training parameters to improve the learning ability of an obtained SR model. 

The remaining organization of this paper is as follows. Section II provides related illustrations of the proposed super-resolution. Section III describes the designed CSRNet. Section IV analyzes the rationality of the proposed method and gives experimental results. Section V summarizes the paper. 

\section{Related Work}
\subsection{Multi-level CNN for image super-resolution}
{Deeper networks can extract more structural information for image super-resolution. In addition, the interactions among different layers can facilitate richer hierarchical information and thus improve the performance of image super-resolution \cite{zhang2018residual}. Thus, multi-level CNNs are a popular means to address the issue of image super-resolution \cite{zhang2018residual}.}

To obtain richer information, Yang et al. \cite{yang2021image} utilized the current layer as inputs of all the layers to extract robust structural features for SISR. Lyn et al. exploited applied blocks to improve the learning ability of a super-resolution method \cite{lyn2020multi}. To improve the efficiency of the obtained super-resolution model, a combination of wavelet and CNN can use low-frequency information to predict high-quality images \cite{zhang2020multi}. To overcome the problem of blurry images from using LR images, fusing multi-scale and multi-level mechanisms into U-Nets can obtain low-frequency information for image super-resolution \cite{han2022multi}. To reduce the number of parameters, asymmetric architectures with residual blocks are used to address blurred edges in super-resolved images \cite{chen2024mffn}. To prevent loss of key information, using channel downscaling and upscaling operations can fuse multi-level information to obtain texture and edge details for image super-resolution \cite{chen2024micu}. To reduce the number of parameters, a hybrid attention and channel separation strategy is presented to mine intra-view information to obtain richer detailed information and textures for SISR \cite{li2024multi}. {To improve robustness of an obtained SR model, Wan et al. proposed a parameter-free attention mechanism to reweight features based on global context adaptively to enable dynamic feature recalibration to achieve an adaptive SISR method
\cite{wan2024swift}. To improve the efficiency of SISR, Park et al. introduced cross-block attention sharing via reusing arcoss networks to mitigate redundancy across hierarchical representations to improve the performance of an SISR model \cite{park2025efficient}. To maintain competitive performance without extra parameters, Gao et al. designed a compact CNN architecture via
 efficient fusion modules to integrate multi-path residual features to improve the effect of restorted images  \cite{gao2023lightweight}.} 
 
According to the description above, it is known that multi-level feature fusion is useful to extract richer structural information. Differing from these methods by primarily refining features within a homogeneous pathway, we use a heterogeneous architecture to simultaneously enhance homologous and heterogeneous structural information for image super-resolution in this paper.

\subsection{CNNs with optimization mechanisms for image super-resolution}
{Deep networks exploit deeper architectures to progressively extract more accurate structural information \cite{tian2024heterogeneou,chen2017deep}. However, the obtained structural information is not representative. Optimization mechanisms can use prior knowledge or some learning strategies to guide a neural network to extract extra information to improve the robustness of the obtained structural information in image super-resolution \cite{tian2021asymmetric}. Given that images are a type of signal, signal processing techniques prove valuable for image-related tasks, e.g., image super-resolution \cite{singh2023using}. Using sparse coding to guide a network is effective for image super-resolution \cite{chen2020survey}. For instance, using sparse coding in a network can improve training efficiency in back-propagation in image super-resolution \cite{liu2016robust}.} {To stabilize training dynamics and improve edge coherence, Zhang et al. employed heat diffusion as a thermodynamic prior to regularize gradient flow and enhance detail recovery for SISR\cite{zhang2024heat}.}

To address unpaired image super-resolution in real scenes, an unsupervised CNN is proposed \cite{wang2021enhanced}. That is, random noise is fed into a GAN to achieve super-resolution images. Then, using a recurrent updating strategy can update random noise to update the texture and structural information of reference images to improve the effect of image super-resolution. To restore more texture information, a dynamic network can extract different information to extract the semantic region in image super-resolution, according to different images \cite{yang2021dynamic}. To address target high-resolution face images, knowledge distillation techniques with facial prior are used to prevent clarity of obtained face images \cite{wang2022propagating}. To overcome the drawback of long-range dependency modeling, the Transformer with 3D-CNN is used to enhance the spatio-spectral relation to improve the visual effect of hyperspectral image super-resolution \cite{ma2023learning}. To capture richer content in hyperspectral image super-resolution, Wang et al. exploited spatial and spectral information to generate more detailed and physically consistent reconstructions for SISR \cite{wang2024single}. Thus, prior knowledge can guide a network to learn more structural information for image super-resolution, which concerns more structural information of a network rather than the optimal solution of training. Differing from the aforementioned studies of relying on external priors or homogeneous optimization, we refer to a novel structural interaction-driven training strategy to improve the learning ability of an obtained SR model in this paper.

\begin{figure*}[!t]
\centering
\includegraphics[width=6.5in]{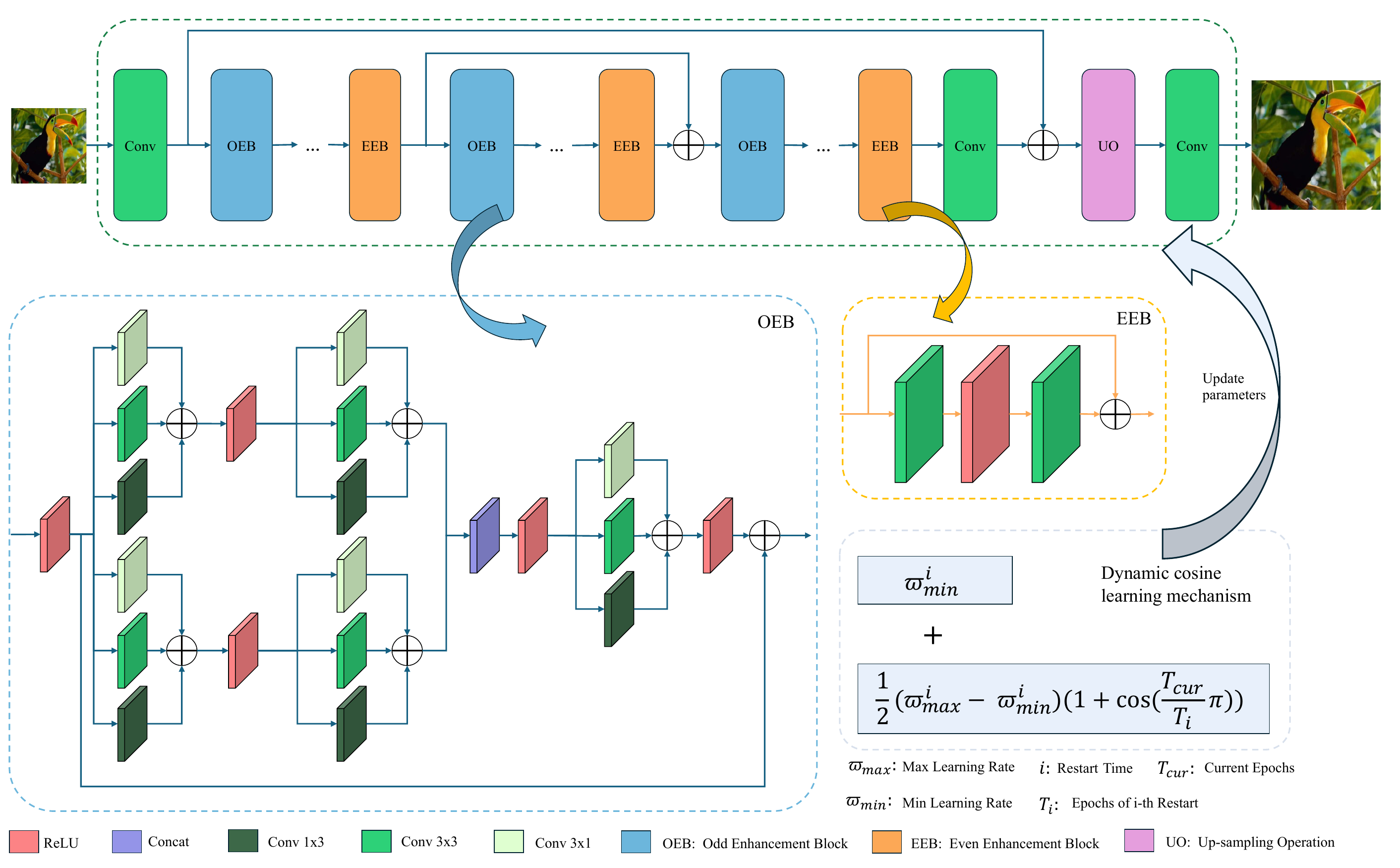}
\caption{Cosine network architecture.}
\label{fig_1}
\end{figure*}
\section{The Proposed Method}
\subsection{Network Architecture}
To enhance the expressive ability of a deep network, we present a cosine network for SISR. The designed CSRNet addresses two key limitations: improving network architecture and optimizing training strategy. Different from previous research on image super-resolution, improving network architecture can design a heterogeneous network architecture to simultaneously enhance the relationship between homologous and heterologous structural information, thereby improving the super-resolution performance. It is composed of convolutional layers (Conv), odd enhancement blocks (OEB), even enhancement blocks (EEB), and an up-sampling operation (UO) of four types to implement efficient performance for image super-resolution. Its visual architecture is shown in Fig. \ref{fig_1}. Conv is set in the 1st, 34th, and 36th layers to extract linear structural information, where the Conv kernel size is $3\times3$. More information is as follows: input channels of 3 and output channels of 64 from the 1st layer, input and output channels of 64 from the 34th layer, input channels of 64 and output channels of 3 from the 36th layer. 16 OEBs, which cascade parallel and serial asymmetric convolutional layers followed by ReLU, are inserted into the odd-numbered layers, i.e., the 2nd, 4th, 6th, \ldots, 28th, 30th, and 32nd layers of CSRNet to enhance relations of homologous and heterologous structural information for image super-resolution. 16 EEBs, each stacking identical convolutions, ReLU, and a skip connection, are placed into the even-numbered layers, i.e., the 3rd, 5th, 7th, \ldots, 31st, 33rd layers of CSRNet to enhance relations of hierarchical structural information for image super-resolution. To extract complementary features, both global and local enhancement operations are employed. Specifically, a global enhancement operation based on residual learning (RL) is applied to the outputs of the 1st and 34th layers, aiming to preserve features from shallow layers and thereby mitigate gradient vanishing or explosion. A local enhancement operation via an RL is used as outputs of the 9th and 21st layers to facilitate diverse structural information for image super-resolution. To obtain high-quality images, a UO is used as the 35th layer of CSRNet to map the predicted low-frequency features into high-frequency features as an input of the 36th layer (the last layer). The last layer reconstructs the final super-resolution images (i.e., high-quality images). The overall procedure is formulated in Eqs.~(\ref{eq:csrnet_1}) and (\ref{eq:csrnet_2}).
{
\begin{equation}
\label{eq:csrnet_1}
\begin{aligned}
O_{\text{CSRNet}} &= \text{CSRNet}(I) \\  
&= C\bigl(UO(C(T) + C(I))\bigr)
\end{aligned}
\end{equation}
}
{
\begin{equation}
\label{eq:csrnet_2}
\begin{aligned}
T &= 6\,\text{EEB}\bigl(\text{OEB}(4\,\text{EEB}(\text{OEB}(C(I)))) \\
  &\quad + 10\,\text{EEB}(\text{OEB}(C(I)))\bigr).
\end{aligned}
\end{equation}
}where $I$ stands for given low-resolution image, $CSRNet$ is used as CSRNet, $C$ denotes a convolutional layer, {$T$ represents the output result after the enhancement blocks for odd and even,} $OEB$ denotes an odd enhancement block, $EEB$ denotes an even enhancement block, $4EEB(OEB)$ stands for four combinations of even enhancement block and odd enhancement block. $6EEB(OEB)$ stands for six combinations of even enhancement block and odd enhancement block. $10EEB(OEB)$ stands for ten combinations of even enhancement block and odd enhancement block. $UO$ denotes an up-sampling operation. + stands for a residual learning operation. $O_{CSRNet}$ denotes a predicted super-resolution image.

\subsection{Loss Function}
L1 converges faster than L2 \cite{zhang2015survey}, thus, the mean absolute error (MAE) \cite{error2016mean} is chosen as the loss function of CSRNet. Its process is shown as follows. Firstly, the low-resolution image $I_i$ is fed into CSRNet to predict the super-resolution image $O_{\text{CSRNet}}^i$. Second, MAE computes the difference between the predicted image $O_{\text{CSRNet}}^i$ and the reference high-resolution image $R_i$ to optimize parameters. The above procedure is formulated in Eq.~\eqref{eq:loss}.

\begin{equation}  
\label{eq:loss}  
\begin{split}
M(\theta) 
&= \frac{1}{N} \sum_{i=1}^{N} \left| \text{CSRNet}(I_i) - R_i \right| \\
&= \frac{1}{N} \sum_{i=1}^{N} \left| O^i_{\text{CSRNet}} - R_i \right|,
\end{split}
\end{equation}
where $M$ is the MAE loss, $\theta$ denotes trainable parameters, and $N$ is the batch size. All parameters are optimized by the Adam optimizer \cite{kingma2014adam} together with the cosine annealing mechanism described in {Section~III.E}.

\subsection{Odd enhancement blocks}
{Each odd enhancement block can cascade parallel and serial asymmetric convolutional layers and ReLU to enhance relations of homologous and heterologous structural information for image super-resolution. It is implemented by two components, i.e., asymmetric convolutional layers and ReLU, where asymmetric convolutional layers are composed of a convolutional layer of $1\times3$, a convolutional layer of  $3\times3$ and a convolutional layer of  $3\times1$ in a parallel way. Firstly, a ReLU is used as the first layer to convert linear information to non-linear information, which acts as a dual network to extract complementary information. Secondly, each sub-network comprises two cascading asymmetric convolutional layers and a ReLU layer acts between the two cascaded asymmetric convolutional layers. Also, two sub-networks are integrated by a concatenation operation as the start of a ReLU. Thirdly, the end of this ReLU is used as an input of asymmetric convolutional layers, which is used as an input of a ReLU. Finally, to improve the expressive ability of this odd enhancement block, an RL is embedded to act between the outputs of the first ReLU and the last ReLU. Specifically, asymmetric convolutional layers are used to extract homologous structural information. Also, ReLU is used to extract heterologous structural information. Besides, parallel and serial architectures can facilitate diverse information. The mentioned techniques can not only extract complementary structural information, but also facilitate the expressive ability of the proposed network for SISR.} We can use Eqs.~\eqref{eq3} and \eqref{eq4} to express the process above. 
\begin{equation}  
    \label{eq3}  
    \begin{aligned}  
        O_{OEB} &= OEB(O_t) \\  
                &= R\left(Cat\left(A\left(R\left(A\left(R\left(A(R(O_t))\right)\right)\right),\right.\right.\right. \\  
                &\quad\quad\quad\left.\left. A\left(R\left(A\left(R(O_t)\right)\right)\right)\right)\right)) + R(O_t).  
    \end{aligned}  
\end{equation}
\begin{equation}
    \label{eq4}
    \begin{array}{l}
        A(t) = C _ { 1 } ( t ) + C ( t ) + C _ { 3 } ( t ),
    \end{array}
\end{equation}
where  $O_t$ and $t$ are used to express temporary inputs. $R$ is a ReLU. $A$ is a combination of asymmetric convolutional layers. $Cat$ is a concatenation operation.  $C_1$ is a convolutional layer of $1\times3$ and $C_3$ is a convolutional layer of $3\times1$.  $O_{OEB}$  is an output of an OEB. 
\subsection{Even enhancement blocks}
An EEB can use stacked equal convolutions, ReLU and a skip connection to enhance relations of hierarchical structural information for image super-resolution. Its implementations via two convolutional layers of $3\times3$, a ReLU and an RL. A ReLU is embedded between two convolutional layers to convert linear information to non-linear information for image super-resolution. A residual learning operation is applied between the start of a convolutional layer and the end of a convolutional layer to strengthen the relations of hierarchical structural information to improve the performance of SISR. The difference between odd enhancement blocks and even enhancement blocks can facilitate more useful structural information to improve the super-resolution effect. The mentioned procedure can be conducted to Eq.~\eqref{eq5}.
\begin{equation}  
    \label{eq5}  
    \begin{split}  
        O_{EEB} &= EEB(O_{t1}) \\  
                &= C(R(C(O_{t1}))) + O_{t1},  
    \end{split}  
\end{equation}
where $O_{t1}$ is an input of an EEB, $O_{EEB}$ is an output of an EEB. 

\subsection{Cosine optimizer}
To optimize parameters, a cosine optimizer is used to dynamically adjust the learning rate to prevent the local minimum of gradient descent and to accelerate the speed of training for image super-resolution \cite{loshchilov10stochastic}. Its implementation can be expressed as Eq.~\eqref{eq:cosine}. 
\begin{equation}
    \label{eq:cosine}
    \begin{array}{l}
    \varpi_t = \varpi_{\text{min}}^i + \frac{1}{2}\left(\varpi_{\text{max}}^i - \varpi_{\text{min}}^i\right) \left(1 + \cos\left(\frac{T_{\text{cur}}}{T_i}\pi\right)\right),
    \end{array}
\end{equation}
where $\varpi_t$ is the current learning rate. $i$ is the number of restarts.  $\varpi_{\text{min}}^i$ is the minimum value of the learning rate of the ith restarting.  $\varpi_{\text{max}}^i$ is the maximum value of the learning rate of the ith restarting.  $T_i$ is the number of the ith restart. $T_{cur}$ is the current epoch. $cos$ is the cosine function. 

\begin{table}[t]
\centering
\caption{Summary of Experimental Settings for CSRNet Training and Evaluation}
\label{tab:exp_settings}
\begin{tabular}{ll}
\toprule
\textbf{Component} & \textbf{Setting} \\
\midrule
Training Dataset & DIV2K (800 images) \\
Validation Dataset & DIV2K (100 images) \\
Test Datasets & Set5, Set14, B100, Urban100 \\
Scaling Factors & $\times2$, $\times3$, $\times4$ \\
Evaluation Metrics & PSNR, SSIM \\
Color Space for Evaluation & Y channel of YCbCr \\
\midrule
Patch Size (HR) & $48 \times 48$ \\
Data Augmentation & Random horizontal flip, 90° rotation \\
Batch Size & 16 \\
Training Epochs & 300 \\
Optimizer & Adam\\
Adam Parameters & $\beta_1 = 0.9$, $\beta_2 = 0.999$, $\epsilon = 10^{-8}$ \\
Initial Learning Rate & $1 \times 10^{-4}$ \\
Learning Rate Schedule & Cosine Annealing with Warm Restarts \\
\midrule
Implementation Framework & PyTorch 1.8.1, Python 3.6 \\
Hardware & 2 $\times$ NVIDIA GeForce RTX 3090 GPUs \\
CUDA Version & 11.1 \\
OS & Ubuntu 18.04 \\
\bottomrule
\end{tabular}
\end{table}

\section{Experimental analysis and results }
\subsection{Datasets}
To fairly evaluate the super-resolution performance of the proposed CSRNet, the public DIV2K \cite{agustsson2016challenge} dataset is used for training. The DIV2K dataset contains 800 training images, 100 validation images, and 100 test images; only the 800 training images are used for model training. Test images are taken from four benchmark datasets: Set5 \cite{bevilacqua2012low}, Set14 \cite{zeyde2012single}, B100 \cite{martin2001database} and Urban 100 (U100) \cite{huang2015single}, with scaling factors of $\times2$, $\times3$ and $\times4$. Evaluation metrics are PSNR \cite{hore2010image} and SSIM \cite{hore2010image}. Test images are converted to the Y channel of YCbCr space to validate the SR performance of CSRNet, following common practice.
\subsection{Parameter settings}
To obtain more training samples, LR and HR pairs are cropped as $48\times48$. Then, these image patches can be augmented by random horizontal flips and 90° rotations to expand the diversity of training samples. Also, $\beta_1$= 0.9, $\beta_2$= 0.999, $\epsilon=10^{-8}$, the learning rate is initialized to $10^{-4}$, the minibatch size is 16, the epoch is 300, and more parameters can be optimized by Adam \cite{kingma2014adam}. {The hyperparameter configuration, including data augmentation and optimization details, is provided in Table~\ref{tab:exp_settings}.}

\subsection{Ablation Study}
Some popular methods, i.e., VDSR \cite{kim2016accurate}, HGSRCNN\cite{tian2022heterogeneous}, and RDN \cite{zhang2018residual} used a scale of $\times2$ to conduct ablation experiments. Thus, to keep fairness, a scale of $\times2$ is chosen to conduct ablation experiments to verify the effectiveness and rationality of the proposed method for SISR in this paper.

It is known that the maximum use of hierarchical information can facilitate the diversity of obtained information in SISR \cite{zhang2018residual}. However, how to obtain representative hierarchical information is essential for SISR \cite{tian2024heterogeneous}. To deal with the problem, scholars tried to design new architectures by increasing network width and depth to enlarge the receptive field for image super-resolution \cite{wang2020deep}. Although that can improve super-resolution performance, it may suffer from challenges in balancing performance, parameter count, and depth/width in image super-resolution. Besides, due to deeper or wider networks, network training difficulty may be increased. To overcome this drawback, a heterogeneous network is designed to extract more useful structural information in this paper. The mentioned network is designed to enhance network architecture and optimize training strategy. 

For the first aspect, odd and even enhancement blocks are designed to obtain complementary homologous structural information to improve the recovery effect of super-resolution images. Also, different components can extract rich heterogeneous information, which is complementary to the obtained homologous information to improve the robustness of the obtained structural information for image super-resolution. Further details of the odd and even enhancement blocks are given below.

Odd enhancement block: Inspired by GoogLeNet \cite{szegedy2015going} and VGG \cite{simonyan2014very}, increasing depth and width can improve the performance of image applications. However, deeper networks may cause gradient vanishing and explosion. Wider networks may cause overfitting. To inherit their advantages, a parallel-and-serial network is designed in an odd enhancement block. The parallel network is combined via two sub-networks. 

The enhancement of each sub-network is implemented in three steps. The first step uses asymmetric convolutions in the sub-network to extract complementary homologous structural information, where asymmetric convolutions are composed of convolutional layers of $3\times1$, $1\times3$, and $3\times3$ can mine salient information in terms of horizontal and vertical directions. Its effectiveness can be verified by ‘A combination of two sub-networks with asymmetric convolutions, three ReLUs and a Conv’ and ‘Three combinations of Conv and ReLU’ in TABLE ~\ref{tab:table1}. Also, two sub-networks are fused by a concatenation operation. The second step uses two cascaded asymmetric convolutions in each sub-network to extract more information. Its effectiveness can be verified by ‘OEB without serial asymmetric convolutions and a residual learning operation’ and ‘A combination of two sub-networks with asymmetric convolutions, three ReLUs and a Conv’. The third step is that a ReLU is set between two asymmetric convolutions to extract heterogeneous structural information. The effectiveness of ReLU can be verified by ‘Three combinations of Conv and ReLU’ and {‘Three Convs’}, where {‘Three Convs’} denotes three stacked convolutional layers. To enhance the expressive ability of the obtained structural information from a wider network, a serial network is used to extract deep structural information. It uses asymmetric convolutions to implement interaction between the parallel and serial branches to facilitate richer homologous information for image super-resolution. Its effectiveness can be verified by ‘OEB without a residual learning operation’ and ‘OEB without serial asymmetric convolutions and a residual learning operation’ in TABLE ~\ref{tab:table1}. To make the obtained information more expressive, ReLU is placed at the ends of the parallel and serial networks. Finally, to prevent long-term dependency problems of a deep network, a residual connection is inserted between the outputs of the two ReLUs. In the odd enhancement block, to further improve the robustness of the obtained structural information for image super-resolution. Its effectiveness can be verified by ‘OEB without a residual learning operation’ and ‘OEB’ in TABLE ~\ref{tab:table1}. The Effectiveness of OEB can be verified via ‘EDSR with EEBs, OEBs’ and ‘EDSR with EEBs’ in TABLE ~\ref{tab:table3}.
\begin{table}[!t]
\centering
\caption{Ablation experimental results of key components on set14 from an odd enhancement block.}
\label{tab:table1}
\renewcommand\arraystretch{1} 
\begin{tabular}{p{6cm} c}
\toprule
Methods ($\times2$) & PSNR (dB) \\
\midrule
{Three Convs} & 12.704 \\
\addlinespace 
Three combinations of Conv and ReLU & 12.948 \\
\addlinespace
A combination of two sub-networks with asymmetric convolutions, three ReLUs and a Conv & 13.865 \\
\addlinespace
OEB without serial asymmetric convolutions and a residual learning operation & 17.729 \\
\addlinespace
OEB without a residual learning operation & 27.881 \\ 
\bottomrule
\end{tabular}
\end{table}

\begin{table}[!t]
\centering
\caption{Ablation experimental results of key components from an even enhancement block on set14.}
\label{tab:table2}
\renewcommand\arraystretch{1}
\begin{tabular}{l c}
\toprule
Methods ($\times2$) & PSNR (dB) \\
\midrule
{Two Convs} & 14.121 \\
EEB without a residual learning operation & 17.653 \\
EEB & 27.468 \\
\bottomrule
\end{tabular}
\end{table}
\begin{table}[htbp]
\centering
\caption{PSNR (dB) of different methods on set14 for $\times2$.}
\label{tab:table3}
\renewcommand\arraystretch{1}
\begin{tabular}{l c}
\toprule
Methods ($\times2$) & PSNR (dB) \\
\midrule
{EDSR} & 34.02 \\
EDSR with EEBs & 34.08 \\
EDSR with EEBs, OEBs & 34.09 \\
CSRNet & 34.12 \\
\bottomrule
\end{tabular}
\end{table}
Even enhancement block: To prevent the naïve effect of OEB, even enhancement blocks are designed to enhance relations of hierarchical structural information for image super-resolution. That is, each EEB utilizes two stacked identical $3\times3$ convolutions to refine the obtained structural information from each OEB. Also, a ReLU is set between two equal convolutions to extract non-linear information for image super-resolution. Their effectiveness is verified by ‘Two Convs’ and ‘EEB without a residual learning operation’ in TABLE ~\ref{tab:table2}. To use hierarchical information, a residual connection is embedded between the start of a Conv and the end of another Conv. Also, its effectiveness is tested via ‘EEB’ and ‘EEB without a residual learning operation’ in TABLE ~\ref{tab:table2}. Besides, the effectiveness of OEB can be verified via ‘EDSR with EEBs, OEBs’ and ‘EDSR with EEBs’ in TABLE ~\ref{tab:table3}. 

\begin{table}[htbp]
\centering
\caption{SISR results of CSRNet with different positions on Set14 for $\times2$.}
\label{tab:table4}
\scalebox{0.82}[0.9]{%
\renewcommand\arraystretch{1.2}
\begin{tabular}{lc}
\toprule
Locations of residual learning operations in CSRNet ($\times2$) & PSNR (dB) \\
\midrule
1,11,21,34 & 34.06 \\
\addlinespace
1,7,21,34 & 34.06 \\
\addlinespace
1,9,21,34 (Our CSRNet) & 34.12 \\
\bottomrule
\end{tabular}%
}
\end{table}
To prevent long-term dependency problems, a two-phase multi-level fusion mechanism is proposed to enhance the expressive ability of the proposed CSRNet. That is, the first phase uses a residual connection to act between the end of the first Conv and the end of the second Conv to enhance homologous information to improve the performance of SISR. The second phase uses a residual connection \cite{li2023stereo} to act between outputs of the 4th and 10th EEBs to facilitate diverse structural information for image super-resolution. Their effectiveness is verified via ‘CSRNet’ and ‘EDSR with EEBs, OEBs’ in TABLE ~\ref{tab:table3}. Also, the choice of acted locations of the two residual learning operations has the following reasons. Firstly, to prevent long-term dependency problems, a residual connection is used as the end of the first layer and the last layer of low-frequency structural information to improve the expressive ability of the obtained super-resolution model, according to VDSR \cite{kim2016accurate}. Differing from VDSR, because heterogeneous information is exclusive, we integrate the output of the first layer and the last layer of low-frequency to fuse homologous information to improve the quality of the obtained images. To quickly extract effective information, we choose the binary search algorithm \cite{lin2019binary,dong2024uafer} to search effective layer for image super-resolution. That is, there are 34 layers in CSRNet to extract low-frequency information. According to binary search, the 17th or 18th layers should be at the end of the second residual learning operation. Thus, the midpoint of 1-17 layers is used as the start of the second residual learning operation, where an output of the ninth layer $(17/2=8.5\approx9)$ is chosen as the start of the second residual learning operation, according to binary search. End of the second residual learning operation is chosen from the 18th-34th layers, where its midpoint $(18+(34-18)/2=26)$ is considered as the end of the second residual learning operation. Taking into account even and odd enhancement blocks, their midpoint $(33/2=16.5)$ is also considered as the end of the second residual learning operation. To make a tradeoff between even enhancement blocks, odd enhancement blocks, and the whole heterogeneous architecture, their midpoint $((16.5+26)/2=21.25\approx21)$ between 16.5 and 26 is chosen as the end of the second residual learning operation. Thus, the 1st, 9th, 21st, and 34th layers are chosen as ends of two residual learning operations. Its effectiveness can be verified by ‘EDSR with EEBs, OEBs, and ‘CSRNet’ in TABLE ~\ref{tab:table3}. 

\begin{table}[!t]
\centering
\caption{Average PSNR result of Cosine with different training strategies on Set14 for $\times2$.}
\label{tab:table5}
\scalebox{0.82}[0.9]{%
\renewcommand\arraystretch{1.2}
\begin{tabular}{lc}
\toprule
Methods ($\times2$) & PSNR (dB) \\
\midrule
Gradient Descent & 33.90 \\
\addlinespace
Cosine & 33.93 \\
\bottomrule
\end{tabular}%
}
\end{table}

\begin{table}[htbp]
\caption{SISR RESULT OF DIFFERENT MODELS ON FOUR DATASETS FOR $\times2$.}
\label{tab:table6}
\renewcommand\arraystretch{1.2} 
\centering
\setlength{\tabcolsep}{3pt} 
\begin{tabular*}{\columnwidth}{@{} l @{\extracolsep{\fill}} cccc @{}}
\toprule
\multirow{2}{*}{Methods ($\times2$)} & Set5 & Set14 & B100 & U100 \\ 
\cmidrule(l){2-5} 
& \multicolumn{4}{c}{PSNR(dB)/SSIM} \\ 
\midrule
Bicubic & 33.66/0.9299 & 30.24/0.8688 & 29.56/0.8431 & 26.88/0.8403 \\
A+ \cite{timofte2015a+} & 36.54/0.9544 & 32.28/0.9056 & 31.21/0.8863 & 29.20/0.8938 \\
SRCNN \cite{dong2015image} & 36.66/0.9542 & 32.42/0.9063 & 31.36/0.8879 & 29.50/0.8946 \\
FSRCNN \cite{dong2016accelerating} & 37.05/0.9560 & 32.66/0.9090 & 31.53/0.8920 & 29.88/0.9020 \\
VDSR \cite{kim2016accurate} & 37.53/0.9587 & 33.03/0.9124 & 31.90/0.8960 & 30.76/0.9140 \\
LapSRN \cite{lai2017deep} & 37.52/0.9591 & 33.08/0.9130 & 31.08/0.8950 & 30.41/0.9101 \\
DRCN \cite{kim2016deeply} & 37.63/0.9588 & 33.04/0.9118 & 31.85/0.8942 & 30.75/0.9133 \\
DRRN \cite{tai2017image} & 37.74/0.9591 & 33.23/0.9136 & 32.05/0.8973 & 31.23/0.9188 \\
CARN \cite{ahn2018fast} & 37.76/0.9590 & 33.52/0.9166 & 32.09/0.8978 & 31.92/0.9256 \\
MemNet \cite{tai2017memnet} & 37.78/0.9597 & 33.28/0.9142 & 32.08/0.8978 & 31.31/0.9195 \\
HGSRCNN \cite{tian2022heterogeneous} & 37.80/0.9591 & 33.56/0.9175 & 32.12/0.8984 & 32.21/0.9292 \\
RDN \cite{zhang2018residual} & 38.24/0.9614 & \textcolor{blue}{34.01/0.9212} & 32.34/0.9017 & \textcolor{blue}{32.89}/0.9353 \\
EDSR \cite{lim2017enhanced} & \textcolor{blue}{38.24/0.9615} & 33.93/0.9203 & \textcolor{blue}{32.35/0.9021} & 32.88/\textcolor{blue}{0.9357} \\
CSRNet & \textcolor{red}{38.29/0.9617} & \textcolor{red}{34.12/0.9216} & \textcolor{red}{32.40/0.9026} & \textcolor{red}{33.14/0.9373} \\
\bottomrule
\end{tabular*}
\end{table}

Effectiveness of the second residual learning operation is compared by ‘1,9,21,34’, ‘1,7,21,34’, and ‘1,11,21,34’ in TABLE ~\ref{tab:table4}. As shown in TABLE ~\ref{tab:table4}, we can see that our method with ‘1,9,21,34’ has obtained a higher PSNR value than that of CSRNet with other locations, which shows the rationality of residual learning locations in terms of location choice. Specifically, ‘1, 11, 21, 34' denotes that a residual operation is applied between ends of the 1st and 34th layers, and a residual operation is applied between ends of the 11th and 21st layers to improve the generalization of the designed CSRNet.
‘1, 7, 21, 34' denotes that a residual operation is applied between ends of the 1st and 34th layers, and a residual operation is applied between ends of the 7th and 21st layers to improve the generalization of the designed CSRNet.
‘1, 9, 21, 34' denotes that a residual operation is performed between ends of the 1st and 34th layers, and a residual operation is performed between ends of the 9th and 21st layers to improve the generalization of the designed CSRNet. 

\begin{table}[!t]
\caption{SISR RESULT OF DIFFERENT MODELS ON FOUR DATASETS FOR $\times3$.}
\label{tab:table7}
\renewcommand\arraystretch{1.2} 
\centering
\setlength{\tabcolsep}{3pt} 
\begin{tabular*}{\columnwidth}{@{} l @{\extracolsep{\fill}} cccc @{}}
\toprule
\multirow{2}{*}{Methods ($\times3$)} & Set5 & Set14 & B100 & U100 \\ 
\cmidrule(l){2-5} 
& \multicolumn{4}{c}{PSNR(dB)/SSIM} \\ 
\midrule
Bicubic & 30.39/0.8682 & 27.55/0.7742 & 27.21/0.7385 & 24.46/0.7349 \\
A+ \cite{timofte2015a+} & 32.58/0.9088 & 29.13/0.8188 & 28.29/0.7835 & 26.03/0.7973 \\
SRCNN \cite{dong2015image} & 32.75/0.9090 & 29.28/0.8209 & 28.41/0.7863 & 26.24/0.7989 \\
FSRCNN \cite{dong2016accelerating} & 33.18/0.9140 & 29.37/0.8240 & 28.53/0.7910 & 26.43/0.8080 \\
VDSR \cite{kim2016accurate} & 33.67/0.9210 & 29.78/0.8320 & 28.83/0.7990 & 27.14/0.8290 \\
LapSRN \cite{lai2017deep} & 33.82/0.9227 & 29.87/0.8320 & 28.82/0.7980 & 27.07/0.8280 \\
DRCN \cite{kim2016deeply} & 33.82/0.9226 & 29.76/0.8311 & 28.80/0.7963 & 27.15/0.8276 \\
DRRN \cite{tai2017image} & 34.03/0.9244 & 29.96/0.8349 & 28.95/0.8004 & 27.53/0.8378 \\
CARN \cite{ahn2018fast} & 34.09/0.9248 & 30.00/0.8350 & 28.96/0.8001 & 27.56/0.8376 \\
MemNet \cite{tai2017memnet} & 34.09/0.9248 & 30.00/0.8350 & 28.96/0.8001 & 27.56/0.8376 \\
HGSRCNN \cite{tian2022heterogeneous} & 34.35/0.9260 & 30.32/0.8413 & 29.09/0.8042 & 28.29/0.8546 \\
RDN \cite{zhang2018residual} & 34.71/0.9296 & 30.57/0.8468 & 29.26/0.8093 & 28.80/0.8653 \\
NSR \cite{fan2020neural} & 34.62/0.9289 & 30.57/0.8475 & 29.26/0.8100 & \textcolor{blue}{28.83/0.8663} \\
EDSR \cite{lim2017enhanced} & \textcolor{blue}{34.76/0.9297} & \textcolor{blue}{30.60/0.8475} & \textcolor{blue}{29.27/0.8101} & 28.82/0.8662 \\
CSRNet & \textcolor{red}{34.85/0.9304} & \textcolor{red}{30.70/0.8490} & \textcolor{red}{29.34/0.8116} & \textcolor{red}{29.12/0.8706} \\
\bottomrule
\end{tabular*}
\end{table}
\begin{table}[!htbp]
\caption{SISR RESULT OF DIFFERENT MODELS ON FOUR DATASETS FOR $\times4$.}
\label{tab:table8}
\renewcommand\arraystretch{1.2} 
\centering
\setlength{\tabcolsep}{3pt}
\begin{tabular*}{\columnwidth}{@{} l @{\extracolsep{\fill}} cccc @{}}
\toprule
\multirow{2}{*}{Methods ($\times4$)} & Set5 & Set14 & B100 & U100 \\ 
\cmidrule(l){2-5} 
& \multicolumn{4}{c}{PSNR(dB)/SSIM} \\ 
\midrule
Bicubic & 28.42/0.8104 & 26.00/0.7027 & 25.96/0.6675 & 23.14/0.6577 \\
A+ \cite{timofte2015a+} & 30.28/0.8603 & 27.32/0.7491 & 26.82/0.7087 & 24.32/0.7183 \\
SRCNN \cite{dong2015image} & 30.48/0.8628 & 27.49/0.7503 & 26.90/0.7101 & 24.52/0.7221 \\
FSRCNN \cite{dong2016accelerating} & 30.72/0.8660 & 27.61/0.7550 & 26.98/0.7150 & 24.62/0.7280 \\
VDSR \cite{kim2016accurate} & 31.35/0.8830 & 28.02/0.7680 & 27.29/0.0726 & 25.18/0.7540 \\
LapSRN \cite{lai2017deep} & 31.54/0.8850 & 28.19/0.7720 & 27.32/0.7270 & 25.21/0.7560 \\
DRCN \cite{kim2016deeply} & 31.53/0.8854 & 28.02/0.7670 & 27.23/0.7233 & 25.14/0.7510 \\
DRRN \cite{tai2017image} & 31.68/0.8888 & 28.21/0.7720 & 27.38/0.7284 & 25.44/0.7638 \\
CARN \cite{ahn2018fast} & 32.13/0.8937 & 28.60/0.7806 & 27.58/0.7349 & 26.07/0.7837 \\
MemNet \cite{tai2017memnet} & 31.74/0.8893 & 28.26/0.7723 & 27.40/0.7281 & 25.50/0.7630 \\
HGSRCNN \cite{tian2022heterogeneous} & 32.13/0.8940 & 28.62/0.7820 & 27.60/0.7363 & 26.27/0.7908 \\
RDN \cite{zhang2018residual} & 32.47/\textcolor{blue}{0.8990} & 28.81/0.7871 & 27.72/0.7419 & 26.61/0.8028 \\
RNAN \cite{zhang2019residual} & 32.49/0.8982 & 28.83/0.7878 & 27.72/0.7421 & 26.61/0.8023 \\
NSR \cite{fan2020neural} & 32.55/0.8987 & 28.79/0.7876 & 27.72/0.7414 & 26.61/0.8025 \\
EDSR \cite{lim2017enhanced} & \textcolor{blue}{32.52}/0.8983 & \textcolor{blue}{28.86/0.7884} & \textcolor{blue}{27.73/0.7430} & \textcolor{blue}{26.65/0.8041} \\
CSRNet & \textcolor{red}{32.69/0.8998} & \textcolor{red}{28.95/0.7899} & \textcolor{red}{27.81/0.7443} & \textcolor{red}{26.92/0.8098} \\
\bottomrule
\end{tabular*}
\end{table}

Cosine optimizer: To prevent local optimum, the cosine optimizer is used to optimize the learning rate \cite{loshchilov10stochastic}. That is, the cosine optimizer can increase $x$ to decrease the cosine value to adjust the learning rate, and its detailed information can be shown in {Section III.E.} Its effectiveness can be verified via ‘Cosine’ and ‘Gradient Descent’, which has a higher PSNR than that of ‘Gradient Descent’ on Set14 in TABLE ~\ref{tab:table5}. 

\begin{figure*}[htbp]
\centering
\includegraphics[height=7.2in]{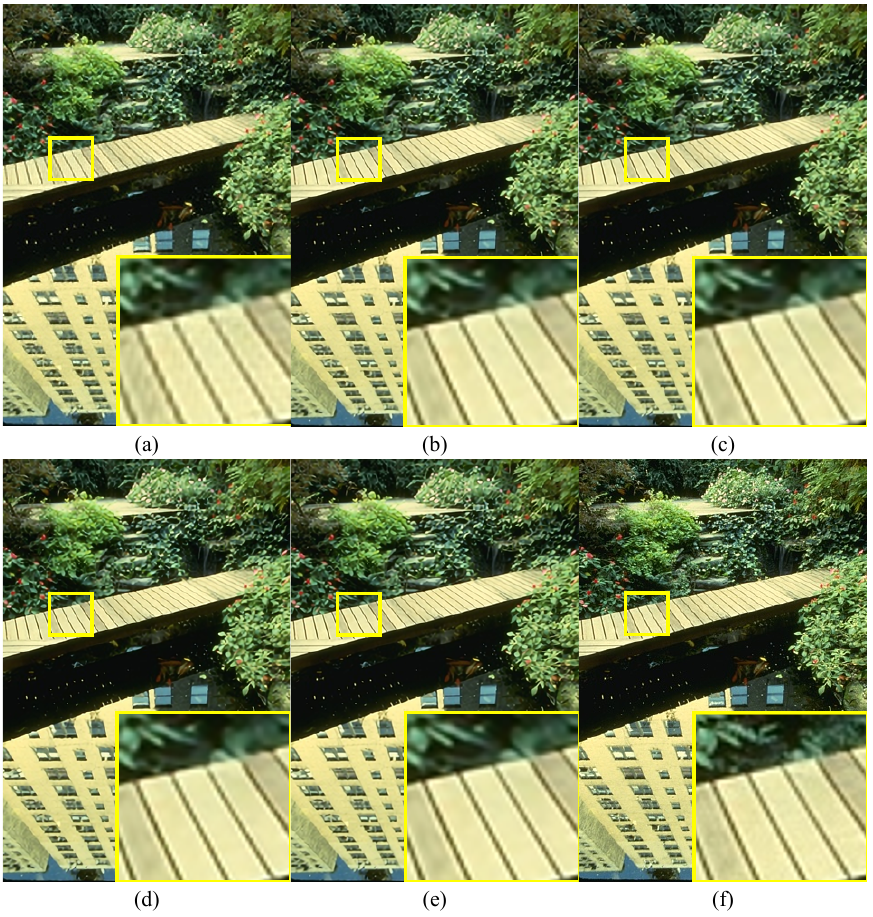}
\caption{Visual results of different methods for ×2 on B100: (a) SRCNN, (b) VDSR, (c) DRCN, (d) CARN, (e) EDSR, and (f) CSRNet (Ours).}
\label{fig_2}
\end{figure*}
\subsection{Experimental results}
To fairly test the performance of our CSRNet for SISR, quantitative and qualitative analyses are used to evaluate our CSRNet. Quantitative analysis is conducted by comparing our CSRNet with A+\cite{timofte2015a+}, a SRCNN\cite{dong2015image}, fast super-resolution (FSRCNN) \cite{dong2016accelerating}, residue context network (RCN) \cite{shi2017structure}, very deep convolutional network (VDSR) \cite{kim2016accurate}, Laplacian pyramid SR network (LapSRN) \cite{lai2017deep}, context-wise network fusion approach (CNF) \cite{ren2017image}, deeply-recursive convolutional SR network (DRCN) \cite{kim2016deeply}, deep recursive residual network (DRRN) \cite{tai2017image}, cascading residual network (CARN) \cite{ahn2018fast}, memory network (MemNet) \cite{tai2017memnet}, multiple degradation convolutional network (SRMDNF) \cite{zhang2018learning}, hierarchical dense recursive network (HDN) \cite{jiang2020hierarchical}, heterogeneous group convolutional network (HGSRCNN) \cite{tian2022heterogeneous}, deep back-projection network (DBPN) \cite{haris2018deep}, residual dense network (RDN) \cite{zhang2018residual}, residual non-local attention network (RNAN) \cite{zhang2019residual}, feedback network (SRFBN) \cite{li2019feedback}, novel sparse representation in deep network (NSR) \cite{fan2020neural} and deep residual network (EDSR) \cite{lim2017enhanced} on four public datasets: Set5 \cite{bevilacqua2012low}, Set14 \cite{zeyde2012single}, B100 \cite{martin2001database} and Urban100 \cite{huang2015single}, with scaling factors $\times2$, $\times3$ and $\times4$ to test PSNR \cite{hore2010image} and SSIM \cite{hore2010image}. Our CSRNet achieves the highest PSNR and SSIM values on both small and large datasets. 

\begin{figure*}[htbp]
\centering
\includegraphics[width=7in]{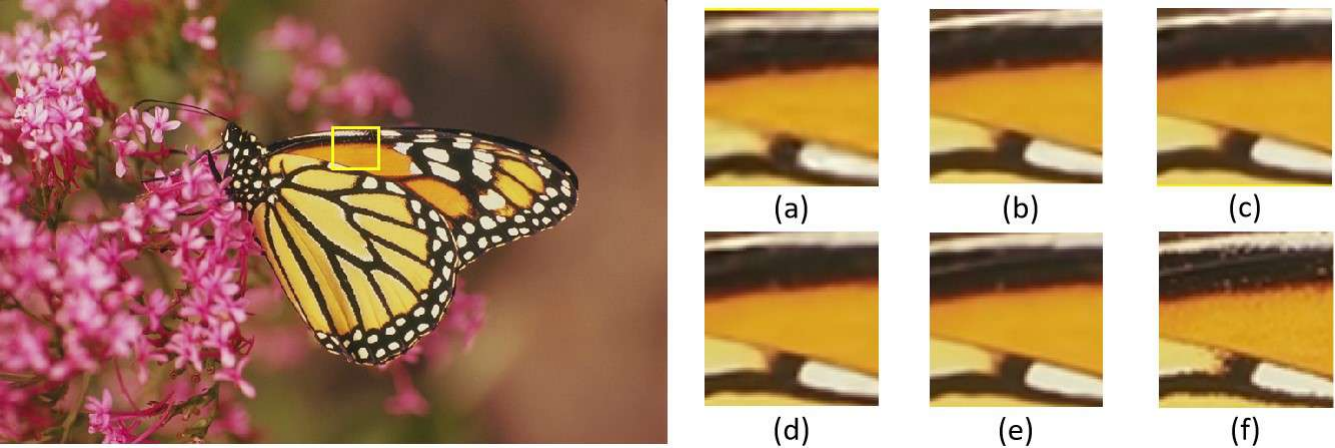}
\caption{Visual results of different methods for ×3 on Set14: (a) SRCNN, (b) VDSR, (c) DRCN, (d) CARN, (e) EDSR, and (f) CSRNet (Ours).}
\label{fig_3}
\end{figure*}
\begin{figure*}[!htbp]
\centering
\includegraphics[width=7.0in]{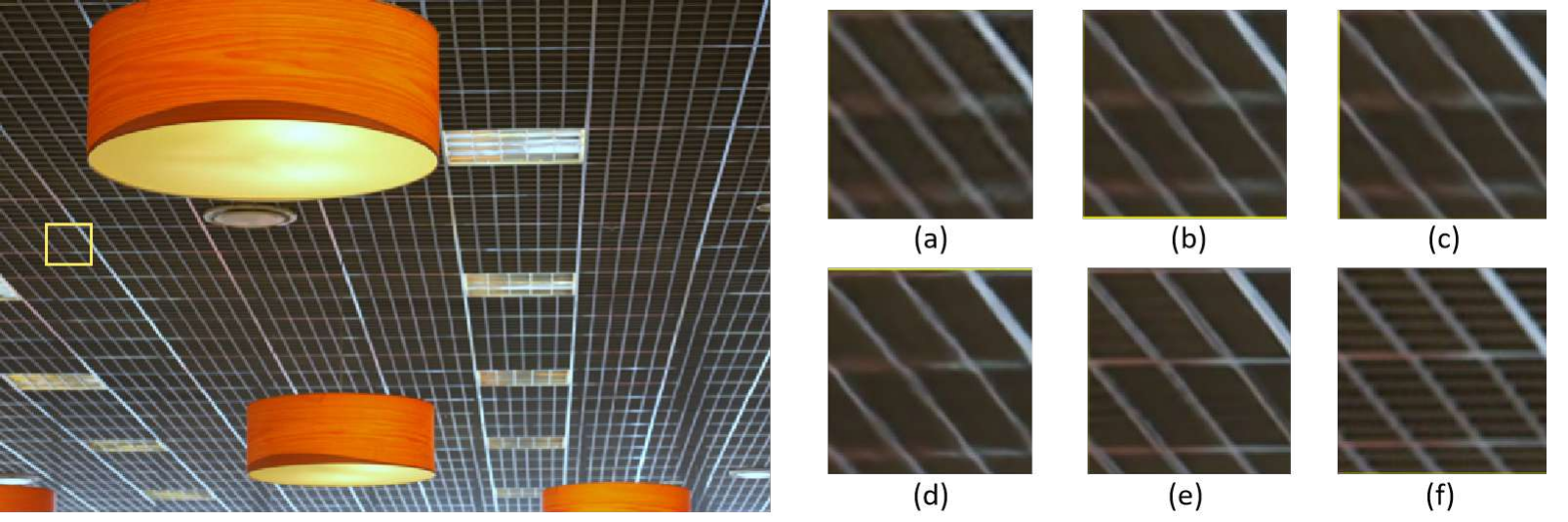}
\caption{Visual results of different methods for ×4 on Urban100: (a) SRCNN, (b) VDSR, (c) DRCN, (d) CARN, (e) EDSR, and (f) CSRNet (Ours).}
\label{fig_4}
\end{figure*}

As shown in TABLE ~\ref{tab:table6}, our CSRNet has an improvement of 0.11 dB in PSNR and 0.0004 in SSIM over the second-best method, RDN on Set14 for $\times2$. As shown in TABLE ~\ref{tab:table7}, our CSRNet has an improvement of 0.29 dB in PSNR and 0.0043 in SSIM over the second-best method RDN, on U100 for $\times3$. As shown in TABLE ~\ref{tab:table8}, our method has obtained an improvement of 0.27dB and SSIM of 0.0057 over that of the second EDSR on U100 for $\times4$. Specifically, the red line denotes the best result and the blue line denotes the second result in Table ~\ref{tab:table6} to ~\ref{tab:table8}. Thus, our CSRNet is very competitive for image super-resolution in quantitative analysis.

We further compare visual results of representative SR methods, including SRCNN, VDSR, DRCN, CARN, and EDSR, as shown in Figs. \labelcref{fig_2,fig_3,fig_4}. In Fig. \ref{fig_2}, CSRNet restores sharper deck lines with clearer structural regularity than those of other methods on an image from the B100 ($\times2$). In Fig. \ref{fig_3}, it preserves finer wing contours and textures of the butterfly than those of other methods on an image from the Set14 ($\times3$). In Fig. \ref{fig_4}, it achieves more faithful structural recovery and sharper textures than competing methods on an image from the Urban100 ($\times4$). Overall, CSRNet confirms the effectiveness of our design in qualitative evaluation.

\section{Conclusion}
In this paper, we present a cosine network for image super-resolution. It depends on improving the network and optimizing the training strategy to implement an efficient performance of image super-resolution. In terms of designing a new network, a heterogeneous network is designed by proposing odd and even enhancement blocks to extract complementary homologous features to improve the quality of recovered images. Enhancing heterogeneous information by combining linear and non-linear structural information can enhance the robustness of the obtained structural information in image super-resolution. Taking into account the local minimum of gradient descent, a cosine annealing mechanism is used to optimize the training process in image super-resolution. Due to complex scenes in the real world, we will model differences of scenes and a deep network to extend our single-scale super-resolution model to an adaptive blind super-resolution model in the future. Also, we will quantify our model to make it suitable for mobile devices.
\bibliographystyle{IEEEtran}
\bibliography{IMSC_AGL}
\end{document}